\documentclass[pmlr]{jmlr}% new name PMLR (Proceedings of Machine Learning)

\usepackage{longtable}% for long tables

 \usepackage{booktabs}
 \usepackage{wrapfig}
 \usepackage{graphicx}
\newcommand{\indep}{\rotatebox[origin=c]{90}{$\models$}}

\usepackage[load-configurations=version-1]{siunitx} % newer version

\makeatletter
\def\set@curr@file#1{\def\@curr@file{#1}} %temp workaround for 2019 latex release
\makeatother

\theorembodyfont{\upshape}
\theoremheaderfont{\scshape}
\theorempostheader{:}
\theoremsep{\newline}

\jmlrvolume{}
\jmlryear{2023}
\title[Understanding Breast Cancer Survival with Causality]{Understanding Breast Cancer Survival: Using Causality and Language Models on Multi-omics Data}

 \author{%
  \Name{Mugariya Farooq$^{*, \S}$}
  \Email{mugariya.farooq@mbzuai.ac.ae}
  \AND
  \Name{Shahad Hardan$^{*, \S}$}
  \Email{Shahad.hardan@mbzuai.ac.ae}
  \AND
  \Name{Aigerim Zhumbhayeva$^\S$} \Email{aigerim.zhumbhayeva@mbzuai.ac.ae}
  \AND
  \Name{Yujia Zheng$^{\ddagger, \S}$}
  \Email{yujiazh@cmu.edu}
  \AND
  \Name{Preslav Nakov$^\S$} \Email{preslav.nakov@mbzuai.ac.ae}
  \AND
  \Name{Kun Zhang$^{\ddagger, \S}$} \Email{kun.zhang@mbzuai.ac.ae}\\
  \addr $^\S$Mohamed Bin Zayed University of Artificial Intelligence, United Arab Emirates \\
  $^\ddagger$Carnegie Mellon University, United States 
}
\begin{document}

\maketitle
\def\thefootnote{*}\footnotetext{These authors contributed equally to this work}
\def\thefootnote{\arabic{footnote}}
\begin{abstract}
The need for more usable and explainable machine learning models in healthcare increases the importance of developing and utilizing causal discovery algorithms, which aim to discover causal relations by analyzing observational data. Explainable approaches aid clinicians and biologists in predicting the prognosis of diseases and suggesting proper treatments. However, very little research has been conducted at the crossroads between causal discovery, genomics, and breast cancer, and we aim to bridge this gap. Moreover, evaluation of causal discovery methods on real data is in general notoriously difficult because ground-truth causal relations are usually unknown, and accordingly, in this paper, we also propose to address the evaluation problem with large language models. In particular, we exploit suitable causal discovery algorithms to investigate how various perturbations in the genome can affect the survival of patients diagnosed with breast cancer. We used three main causal discovery algorithms: PC, Greedy Equivalence Search (GES), and a Generalized Precision Matrix-based one. We experiment with a subset of The Cancer Genome Atlas, which contains information about mutations, copy number variations, protein levels, and gene expressions for 705 breast cancer patients. Our findings reveal important factors related to the vital status of patients using causal discovery algorithms. However, the reliability of these results remains a concern in the medical domain. Accordingly, as another contribution of the work, the results are validated through language models trained on biomedical literature, such as BlueBERT and other large language models trained on medical corpora. Our results profess proper utilization of causal discovery algorithms and language models for revealing reliable causal relations for clinical applications.
\end{abstract}

\section{Introduction}

The application of deep learning (DL) and machine learning (ML) in biomedical sciences paves a new way of understanding the underlying causes of various fatal diseases, including cancer. Since cancer is a multi-factorial disease, different types of data are used to predict or understand its outcomes using various ML approaches. Despite the vast amount of research on the diagnosis and prognosis of breast cancer, it remains the second leading cause of cancer death for women.\footnote{https://www.cdc.gov/cancer/dcpc} Throughout the disease progression, alterations in the genes, such as perturbations in the gene structure, function, or expression, have a significant impact. These perturbations or mutations tend to skew the normal cellular pathways, consequently changing the normal functioning cell to a cancer cell \citep{hanahan2000hallmarks}. 

Our work focuses on two different types of breast cancer: invasive lobular carcinoma (ILC) and invasive ductal carcinoma (IDC). There is a lack of genomic studies that investigate the underlying biological causes of ILC, as the focus is higher on IDC. However, in clinical practice, ILC patients may not show any symptoms at first, and the cancerous areas are difficult to spot on mammograms \citep{dataset}. Multi-omics data can be used to better understand the progression of ILC and the underlying causes of oncogenesis.

Aiming to understand the factors affecting the survival of breast cancer patients, ML approaches can be applied to discover the underlying patterns in gene alterations. Causality is a fundamental notion in science and plays an important role in explanation, prediction under interventions, and decision-making \citep{pearl2009causality}. In contrast to plenty of other ML areas, causal discovery (CD) \citep{spirtes} aims to estimate causal relations among the variables, often represented by Directed Acyclic Graph (DAG) rather than directly making passive predictions.\footnote{In this work, we assume the causal relationship follows a DAG, i.e., there is no feedback loop. Estimating cyclic causal models is more complicated and figuring out whether cyclic models are more appropriate is one line of our future research.} The high interpretability that it provides is especially beneficial in the biomedical field as it aids decision-making and the comprehension of the analysis given by ML models. There are two major search strategies in causal discovery: score-based and constraint-based. Score-based methods, such as Greedy Equivalence Search (GES) \citep{chickering2002optimal} and Fast GES \citep{fges}, select the causal graph based on the score assigned to each candidate graph. Constraint-based methods find causal relationships based on conditional independence constraints discovered from data. Two examples of constraint-based methods are the PC algorithm and the Fast Causal Inference (FCI) \citep{spirtes}. Another recent approach could readily deal with mixed continuous and discrete data types without strong assumptions on the functional relations between the variables \citep{zheng2023generalized}, producing a Generalized Precision Matrix (GPM) to analyze the conditional independence structure in the data, which can then be refined to produce information about the causal structure. The outcome of the aforementioned methods is a DAG or a set of DAGs (often known as an equivalence class) that shows the existence and the nature of the relationship between any two variables. 

Moreover, generally speaking, there is a challenge to the authenticity of the claims made by CD methods due to the absence of ground truth to validate the results obtained from these methods. The ground-truth causal relations are often unknown in real problems \citep{Tu2019NeuropathicPD}. Researchers that use CD in application fields often rely on domain expertise to examine the outcomes. Interestingly, thanks to the developments of Large Language Models (LLMs), which are pretrained on large medical corpora, can actually help in the task of validation of the results from CD methods: they automatically extract relevant information from the literature. Our approach uses state-of-the-art Natural Language Processing (NLP) architectures that help in validating the claims for further authentication by biologists. The usage of NLP methods reduces the cost and the effort spent on annotating data by specialists by giving them a smaller refined set to work with or verifying novel discoveries in the medical field. Several approaches, such as perplexity and masked language modeling, can be leveraged for validation tasks.

To the best of our knowledge, our work is the first that leverages CD methods to understand the factors affecting the survival of breast cancer patients, with LLMs to verify the findings. Our method aims to find and validate the influences of different alterations in genes on vital status.
The contributions of this work can be summarized as follows:
\begin{itemize}
    \item Unlike plenty of breast cancer ML studies, we leverage multi-omics data to unravel patterns concerning the survival of patients through suitable causal discovery methods. The usage of multi-omics data in research is a relatively new and powerful approach that considers multiple levels of biology.
    \item We strategically dealt with a dataset with mixed data types, which is challenging in the case of causality while being cognizant of the established mathematical assumptions.
    \item We propose a novel approach to validate the claims made by CD methods using state-of-the-art NLP models as a way to filter the most relevant claims out of numerous claims made by the models. We believe this validation approach will have direct implications in other application domains.
\end{itemize}
\subsection*{Generalizable Insights about Machine Learning in the Context of Healthcare}
For machine learning models to be embedded into the healthcare system, we should make them explainable enough to allow effective deployment by clinical practitioners. Unlike deep learning models, causality enables the investigation of causes and effects of the different variables in the multi-omics data. In our approach, we leverage CD methods properly, complemented with validation approaches based on LLMs, for understanding the factors affecting the survival of breast cancer patients. Our research supports the usage of data with mixed types that are available in real-life scenarios. %Since there are no concrete methods to validate the claims made by causality, we propose using language models as a way to automatically verify the results. 
Our evaluation approach helps medical practitioners to re-verify smaller subsets of filtered claims hence expediting the verification process. Overall, our study facilitates the adoption of computational causal approaches for healthcare data for both clinical practitioners and machine learning experts.  
% \begin{itemize}
%     \item The usage of CD approaches to identify the intricate causal relationships that occur in the genome of an individual suffering from a particular disease
%     \item Usage of the results to help in the development of personalized treatment plans
%     \item The existence of mixed data types in real-world scenarios and the methods to deal with it
%     \item Usage of reliable and appropriate LLMs for validation to decrease the annotation/validation cost 
%     \item Refined set of claims to be verified by medical professionals
% \end{itemize}
\section{Related Work}\label{RL}
The causal discovery field gains more focus from a theoretical perspective and less so from applications. The potential to apply CD methods in the biomedical sciences is large, but it is still not highly leveraged. The high interpretability provided by causality leads to more reliable decision-making approaches and high-quality intervention procedures for precision medicine.

In genetics, \cite{cgauge} used a dataset from the UK Biobank to develop their framework. The approach combined both causal analysis and Mendelian randomization, which attained the true relationships between features and reduced the false positive rate by 30\%. Another medical application discovered the leading features of Alzheimer's disease using FGES and FCI \citep{alz}. The study included a ``gold standard'' graph that was used as ground truth to evaluate the accuracy of the CD graphs. The results showed the robustness of FGES compared to FCI. 

At the same time, there are relatively few investigations attempting to discover the underlying causes of cancer with CD. \cite{crisp} developed a causal research and inference platform to be implemented on multi-omics data for oncology problems. The study combined six causal discovery approaches into one to provide a more accurate insight into the biological problem. Moreover, \cite{tumor_fci} examined somatic genome alterations (SAGs) to build a causal procedure that found the alterations that are closely related to the tumors. Their approach used The Cancer Genome Atlas and explained the effect of SAGs on the adoption of disease mechanisms in a patient's body. Another study applied CD to understand gene regulation behind head-and-neck carcinoma, but with a model that accounts for missing data \citep{incomplete}. The motivation came from the incorrect assumption of CD methods presuming complete data, and thus, authors connected CD with multiple imputations according to Rubin's rule. They attempted to put less weight on some strong assumptions that CD takes into consideration as they found it led to higher robustness in the outcomes.

To the best of our knowledge, perplexity and masked language modeling have not been used to verify complex biomedical claims, leading to a dearth of literature in this domain. However, recent developments in NLP, especially in language models, motivate using these methods. Measuring the perplexity score of a hypothesis can give us a basic understanding of the authenticity of the claim. As postulated by \cite{DBLP:journals/corr/abs-2006-04666}, compared to the verified claims, any unverified claims or misinformation would have higher perplexity (degree of falseness). Moreover, much research has been done on fine-tuning language models like BERT or its variants, such as SciBERT \citep{DBLP:journals/corr/abs-1903-10676} or Bio-BERT \citep{10.1093/bioinformatics/btz682} for various NLP tasks in the bio-medical domain. 

Additionally, \cite{https://doi.org/10.48550/arxiv.1909.01066} proposed that language models can be used as knowledge bases with the help of underlying relational information intrinsic to the training data. The recent improvement in many challenging NLP tasks can be credited to the focused research on developing task-agnostic architectures. However, there is still the need for task-specific datasets, which was the motivation behind the research by \cite{DBLP:journals/corr/abs-2005-14165}. Recent research in training GPT models on biomedical data has proven to be successful for various tasks. So far, language models have been used for inference in NLP tasks; however, interestingly, as shown in this paper, they can be leveraged to verify causality claims or hypotheses.

\section{Methods and Materials}\label{methods_materials}
\begin{figure*}[ht]
    \centering
    \includegraphics[width=0.9\textwidth]{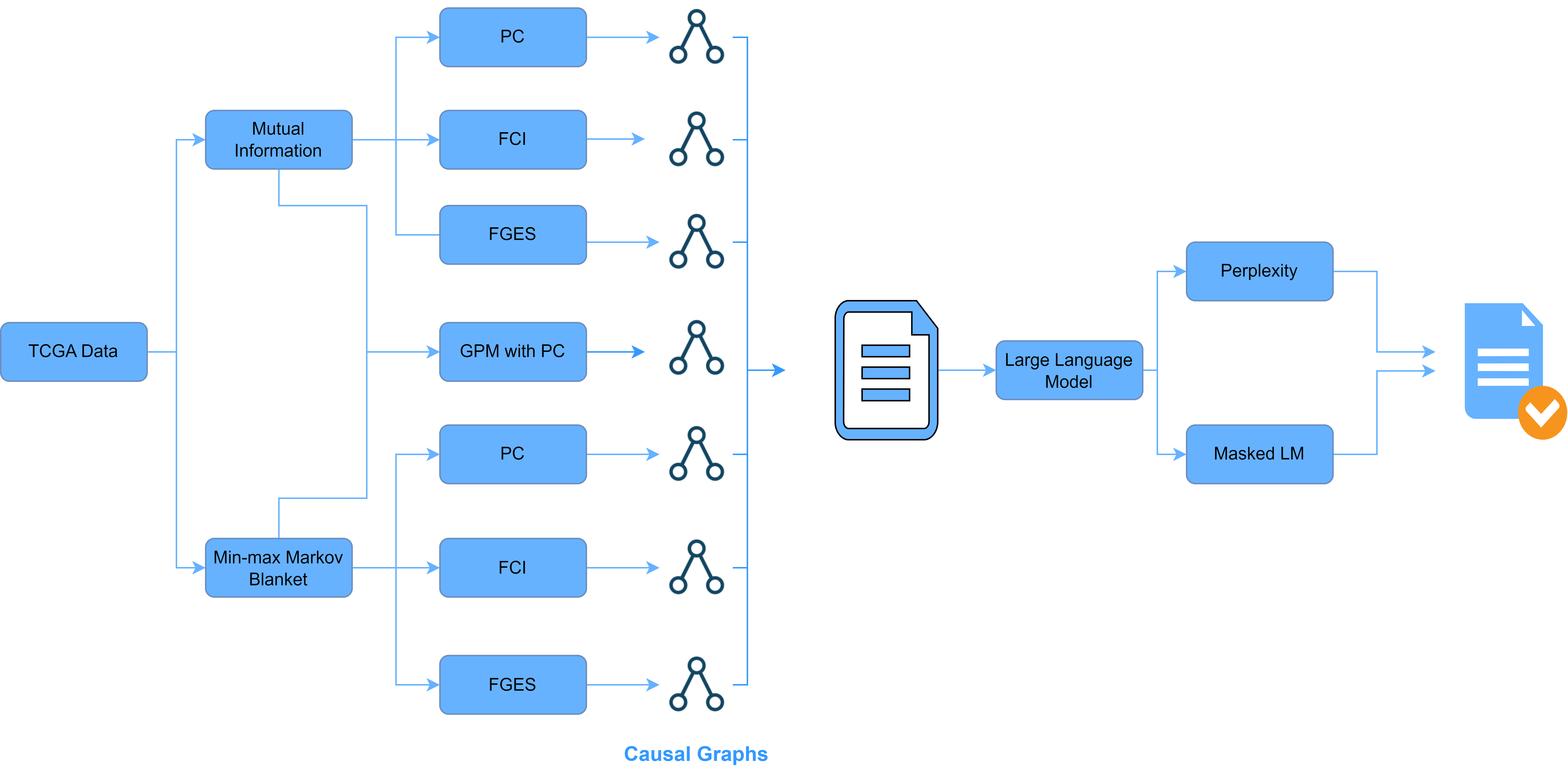}
    \caption{Overall flow of our approach. Due to the size of the relevant TCGA multi-omics data, feature selection needs to be applied first to acquire a smaller subset. Then, four CD methods: PC, FCI, GPM, and GES (or its variant, FGES) were utilized. The CD methods produced causality claims related to the factors that affect the survival of patients. These relationships were then validated using language models through two approaches: perplexity and masked language modeling.}
    \label{overall_arch}
\end{figure*}

In this work, we started with dataset exploration, followed by applying feature selection methods to reduce the number of features as it was initially large in the raw data. Once we had the selected subset, we applied causal discovery methods using the appropriate statistical tests. The overall flow of the method is illustrated in Figure~\ref{overall_arch}.

\subsection{Dataset}
The dataset used is a subset of The Cancer Genome Atlas (TCGA) Breast Cancer dataset \citep{dataset}. It contains records of 705 patients, of which 490 have IDC, 127 have ILC, and 88 have both types.\footnote{www.kaggle.com/code/samdemharter/multi-omics-integration-with-the-qlattice/data} It includes 1,936 features belonging to four main types of variables: copy number variations (represented as \emph{cn}), somatic mutations (represented as \emph{mu}), gene expression (represented as \emph{rs}), and protein levels (represented as \emph{pp}). There are 860 genes with records of copy number variations, which refer to the number of copies in each gene cell. They are represented as categorical variables calculated using the Gistic score. There are records of somatic mutations for 249 genes. Mutation variables are categorical and refer to whether a gene has been mutated or not. Additionally, the dataset contains gene expressions measured by RNA sequencing for 604 genes. The protein levels are also quantified for 223 genes. The gene expression and protein level variables are continuous. All genes in the dataset are represented by their Hugo symbols. Finally, the target variable is the vital status, which refers to whether the breast cancer patient survived or not. Out of the 705 patients, 611 survived and 94 died. We will examine the relationship between the vital status and the four aforementioned types of features. The aim is to understand the features causing the vital status and how the different features affect each other.

\subsection{Data Exploration}
\begin{figure}
         \centering
         \includegraphics[width=0.9\textwidth]{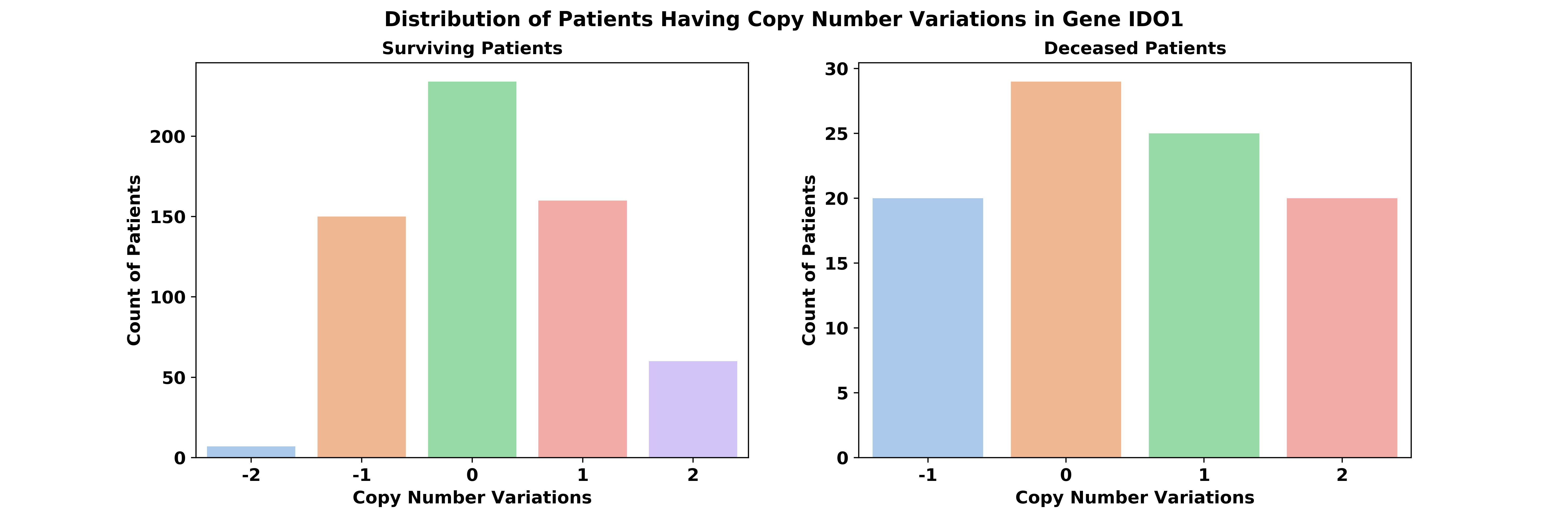}
         \caption{The patient distribution difference across the two vital status categories regarding the copy number variations in gene IDO1.}
         \label{fig:dist_cn_IDO1}
\end{figure}

Since the main outcome of our study is to understand the factors affecting the survival of patients, we visualized the distribution changes of specific variables across two categories of patients: survived and deceased. Regarding copy number variations, Figure~\ref{fig:dist_cn_IDO1} shows their distribution in the gene IDO1, where there is a difference in the distribution and the number of categories present for surviving and deceased patients. Another example of the case of the gene TGFRB3 is shown in Figure~\ref{fig:dist_cn_TGFBR3} in Appendix \ref{app-sect:Data}. Also, the examination of the distribution of gene mutations and the two types of continuous variables can be found in Appendix \ref{app-sect:Data}.

\subsection{Feature Selection}
As the dataset includes over 1,900 variables, there is a need to perform feature selection to reduce the complexity of the causal graphs and to improve the visualization. Our feature selection approach is based on two methods: max-min Markov blanket (MMMB) \citep{mmmb} and mutual information (MI) \citep{Cover1991ElementsOI}, chosen depending on the type of data used. MMMB discovers the local structure by finding the minimal feature subset that consists of parents, children, and parents of the children of a target variable. In \citep{mmmb}, the authors explain that MMMB is based on another CD method, Max-Min parents and children (MMPC), which is able to only find parents and children of a target variable. Therefore, MMPC should be applied before adding the spouses of the target variable. The parents and the children of the target variable are combined with the parents and children of the children's parents and children, building a set called candidate Markov blanket (CMB). After this, a filtering approach is adopted to remove the false positives detected by the additional combinations of variables coming from the different CMBs. It has been demonstrated that MMMB performs well with large datasets and outperforms other Markov blanket methods such as incremental association Markov blanket (IAMB) \citep{mmmb}. For our case, we used MMMB for discrete data with the independence test being multinomial logistic regression, setting the significance level at 0.05. The Markov blanket was derived using the R package \texttt{MXM}.

Due to the limitations of applying MMMB to mixed data, we applied supervised feature selection to choose variables that are related to the target variable (vital status). In this regard, MI feature selection finds a value that is zero if the variables are independent and non-negative if they are dependent, with higher values referring to higher dependency. For this method, we used the Python library \texttt{sklearn}, which allows specifying the discrete variables to obtain accurate results. Upon implementation, we noticed that the variables extracted from MI feature selection were all continuous, which is beneficial to study the relationship between continuous variables in the dataset using a smaller sample size. 

Aiming to produce informative and clear graphs, we only extract 10 variables (including vital status) from the outcome of both feature selection methods. We will then use this subset of features, instead of the whole set of available features, for causal discovery.  It is interesting and important that applying suitable causal discovery methods to the Markov blanket of the target variable, together with the target variable (vital status), is able to identify the parents and children of the target variable \citep{Gao2015LocalCD}. The different CD methods applied to such a subset of the features are introduced in the following section.

\subsection{Causal Discovery Methods}
In our analysis, we used the methods that are broadly applied in learning causal graphs from the data. They can be divided into constraint-based and score-based algorithms. Constraint-based methods find the conditional independencies in the dataset and consequently produce a DAG or a set of DAGs, corresponding to a Markov equivalence class and represented by a pattern, to satisfy those conditional independence constraints \citep{spirtes}. On the other hand, score-based methods rely on finding the optimal graph $\mathcal{G}$ that maximizes a properly defined score of the data given this graph $\mathcal{G}$. The graphs were obtained using \texttt{Tetrad 6.9.0}. In this section, we explain the theory behind four CD approaches: PC, GES, FGES, and GPM. GPM is a nonparametric method that estimates the Markov network which can be used to create a causal structure.

\subsubsection{PC} The PC algorithm \citep{spirtes} assumes the Markov condition and the faithfulness assumption, and its causal discovery results have been shown to be correct in the large sample limit if there are no latent confounders (an unobserved direct common cause of two measured variables). PC starts with a completely undirected graph and deletes the edges based on conditional independence tests by:
\begin{itemize}
    \item Testing all pairs of nodes for marginal independence and deleting the edge if they are marginally independent.  
    \item Testing conditional independence between the remaining adjacent pairs of nodes ($A$, $B$), given any other single node with an edge connected to either of them. If there is any node ($C$) such that $A \indep B | C$, the edge between $A$ and $B$ is removed and node $C$ is saved as a separation set.
    \item Repeating the same procedure of the independence test by increasing the size of the conditioning set one at a time until there are no more adjacent pairs ($A, B$) such that all variables in the conditioning set are adjacent to $A$ or all adjacent to $B$. In the end, we get the skeleton, where all edges are undirected.
\end{itemize}   

Once we have the undirected graph, we start by finding V-structures. For the set of three variables $(A, B, C)$, where only one pair ($A, C$) is not adjacent and other pairs ($A, B$), ($B, C$) are adjacent, orient the edges $A$ - $B$ - $C$ as $A \rightarrow B \leftarrow C$, based on the information saved in the conditioning sets. The next step is orientation propagation. If there is an adjacent node ($D$) to node ($B$) in the V-structure ($A \rightarrow B \leftarrow C$), we form a Y-structure by directing the edge from $B$ to $D$. Finally, we produce the equivalence class that describes the conditional independence information in the data. The edges in the graph can be either undirected or directed \citep{review_cd_kun,PC}. If all DAGs in the equivalence class have the same direction for a particular edge, that edge is directed; otherwise, it is undirected.

\subsubsection{GES and FGES}
GES is one of the score-based methods that starts with an empty graph and adds one edge at a time according to improvements in the score. Then, the resulting Markov equivalence class is formed. This is done until no more improvements in the score can be made when adding an edge. In the second stage, the algorithm goes backward by removing edges until no more improvements in the score can be made. Fast GES (FGES) is a modification of GES that uses parallelization to make the algorithm faster. It assumes a penalty for the score and a weaker version of the faithfulness assumption \citep{fges}. 

\subsubsection{Generalized Precision Matrix}
The work in \citep{zheng2023generalized} proposes using a generalized precision matrix to construct a Markov network. The method aims to address some of the limitations of the previously mentioned algorithms; for instance, 1) the probability measure is assumed to be from a certain family, 2) a CD method can either handle discrete or continuous variables, and 3) having restrictions on the differentiability and the cardinality of the continuous and discrete variables, respectively. Since we have mixed data types in the dataset and some causal influences can be complex in nature, we experiment with GPM with the features chosen from the feature selection approaches. This approach produces a Markov network, which is further refined by the PC algorithm to produce an equivalence class.
\subsection{Statistical Tests}
\label{section:statistical tests}
An essential part of the causal discovery is the statistical tests used for evaluating the relationships between the data. Constraint-based methods, such as PC and FCI, require conditional independence tests for implementation. On the other hand, score-based methods, such as GES and FGES, use scoring methods for model selection of the entire DAG.

\begin{table}[t]
  \centering 
  \caption {Comparison of perplexity scores for different models on the same claim where the claim column represents the direct association of the variables with vital status in the graphs.}
  \resizebox{\columnwidth}{!}{
  \begin{tabular}{llll}
  \toprule
    \textbf{Model} & \textbf{Claim} & \textbf{Perplexity} \\
    \midrule
GPT-2 & ``Mutation in gene UBR4  is related to the survival in cancer" & 76.68\\
GPT-2 & ``Mutation in gene UBR4 is not related to the survival in cancer" & 39.65\\
SciBERT & ``Mutation in gene UBR4  is related to the survival in cancer" & 110.87\\
SciBERT & ``Mutation in gene UBR4 is not related to the survival in cancer" & 167.8\\
BlueBERT& ``Mutation in gene UBR4  is related to the survival in cancer" & 31.01\\
BlueBERT& ``Mutation in gene UBR4 is not related to the survival in cancer" & 2502.9\\
    \bottomrule
  \end{tabular}}
  \label{perp} 
\end{table}

The conditional independence tests adopted for the constraint-based methods are the conditional Gaussian likelihood ratio test, the Chi-square test, and the randomized conditional independence test (RCIT) \citep{rcit}. The conditional Gaussian likelihood ratio test uses a mixture of continuous and discrete variables \citep{scoring_bayesian}. As for the Chi-square test, it is used for testing the independence of categorical variables. RCIT is an approximation for the kernel conditional independence \citep{kci} that speeds up the CD methods and imposes fewer assumptions about the data. Since our dataset is of mixed types, we used the Chi-square test when we exclude the continuous variables and the conditional Gaussian likelihood ratio test and RCIT when we consider a mixed subset.

The score-based methods use multiple scores including discrete BIC \citep{scoring_bayesian} and the conditional Gaussian BIC score \citep{Andrews2019LearningHD}. The discrete BIC test is used only when all variables in the dataset are categorical and are based on a modification to the original BIC score function. The conditional Gaussian BIC score is adopted when the dataset includes discrete and Gaussian variables and is computed under the conditional Gaussian assumption \citep{scoring_bayesian}. Similar to the constraint-based methods, we used the discrete BIC test when the continuous variables are excluded, and the conditional Gaussian BIC score when both types are included. However, these score-based methods are built on some hard assumptions based on the distribution of data in question and the underlying causal mechanisms. Thus, we delved into other forms of score functions for CD named generalized score functions \citep{huang2018generalized}. The score function we used is the generalized score with cross-validation (CV) that calculates the local score where the score setting used is negative k-fold cross-validated log-likelihood. In our code, we used the \texttt{causal-learn} package to use the generalized CV score.

\section{Experiments}\label{exps}
\begin{figure}[ht]
    \begin{minipage}[t]{0.47\textwidth}
        \includegraphics[width=\textwidth]{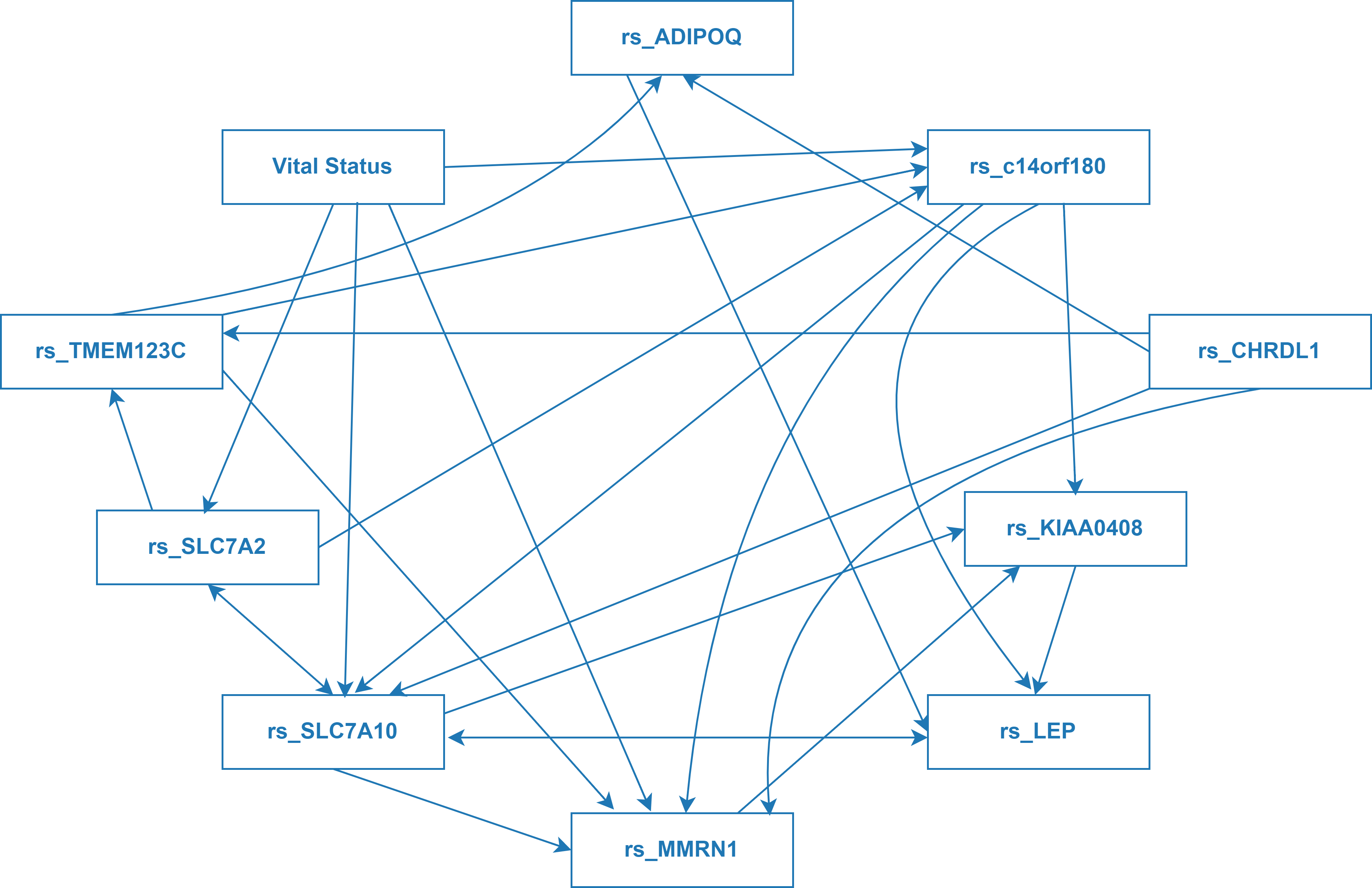}
        \caption{Graph of FGES using the conditional Gaussian BIC score on the mixed data from MI selection.}
        \label{fges_bic_mixed_mi}
    \end{minipage}
    \hfill
    \begin{minipage}[t]{0.41\textwidth}
    \centering
        \includegraphics[width=\textwidth]{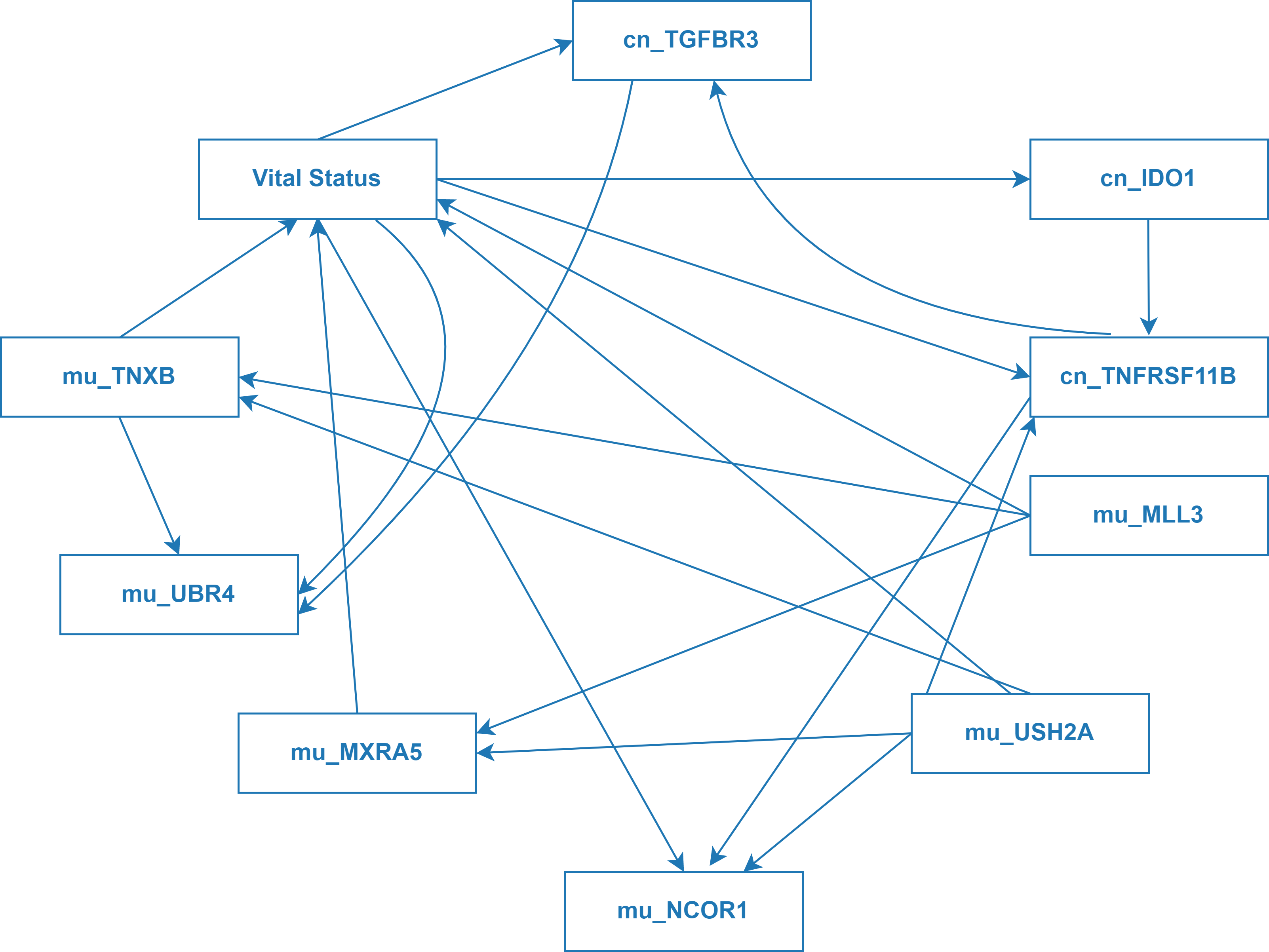}
        \caption{Graph of GES using the generalized cross-validation score on the categorical data from MMMB.}
        \label{ges_mmmmb_cv}
    \end{minipage}
\end{figure}

\begin{figure}[ht]
    \centering
    \begin{minipage}[t]{0.47\textwidth}
        \centering
        \includegraphics[width=\textwidth]{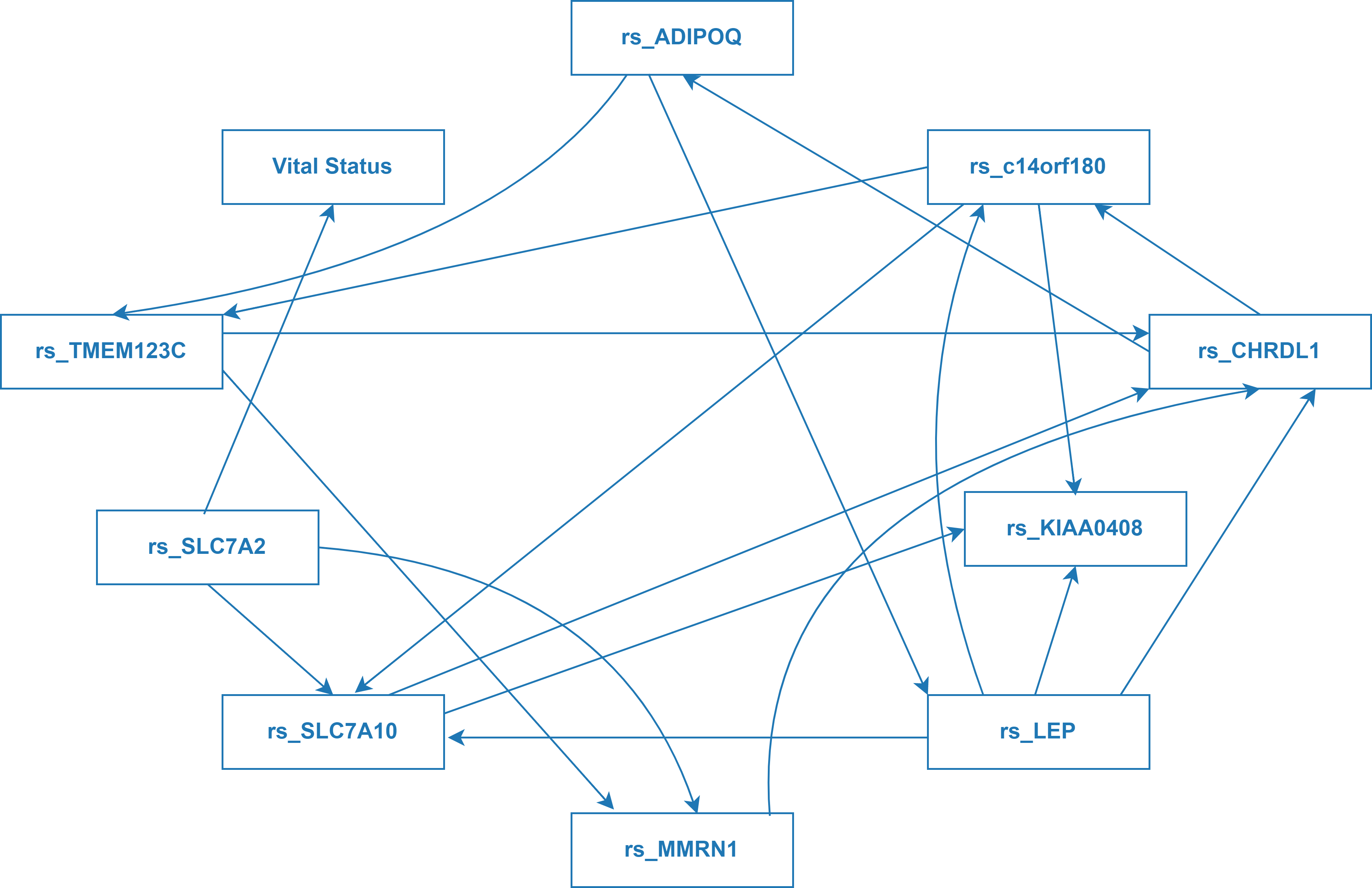}
        \caption{Graph of PC using the conditional Gaussian likelihood ratio test on the mixed data from MI selection.}
        \label{pc_cond_gauss_mi}
    \end{minipage}
    \hfill
    \begin{minipage}[t]{0.47\textwidth}
        \centering
        \includegraphics[width=\textwidth]{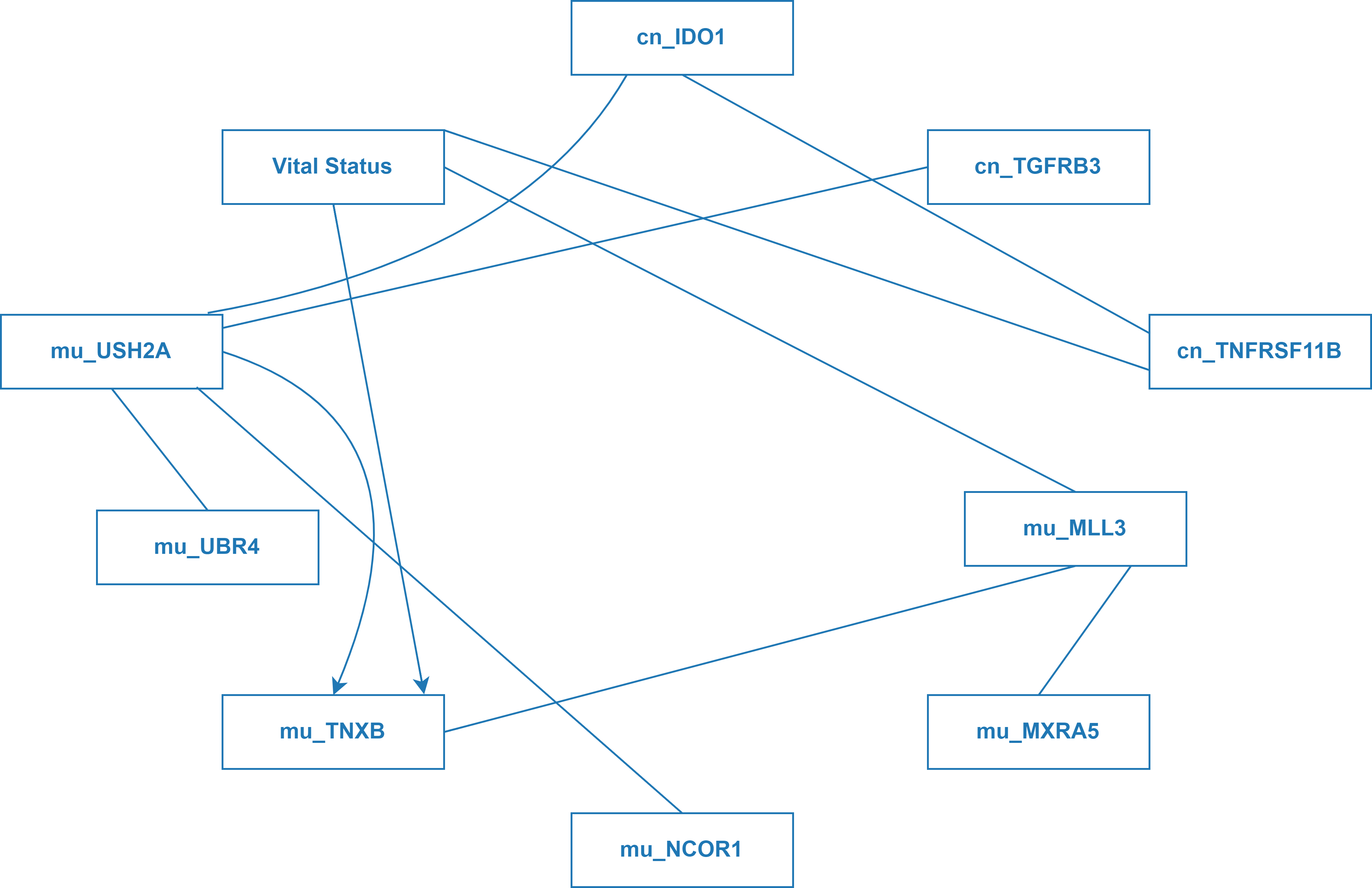}
        \caption{Graph of PC using the Chi-square test on the categorical data from MMMB.}
        \label{pc_chi_categorical}
    \end{minipage}
\end{figure}

\begin{figure}[ht]
    \centering
    \includegraphics[width=0.7\textwidth]{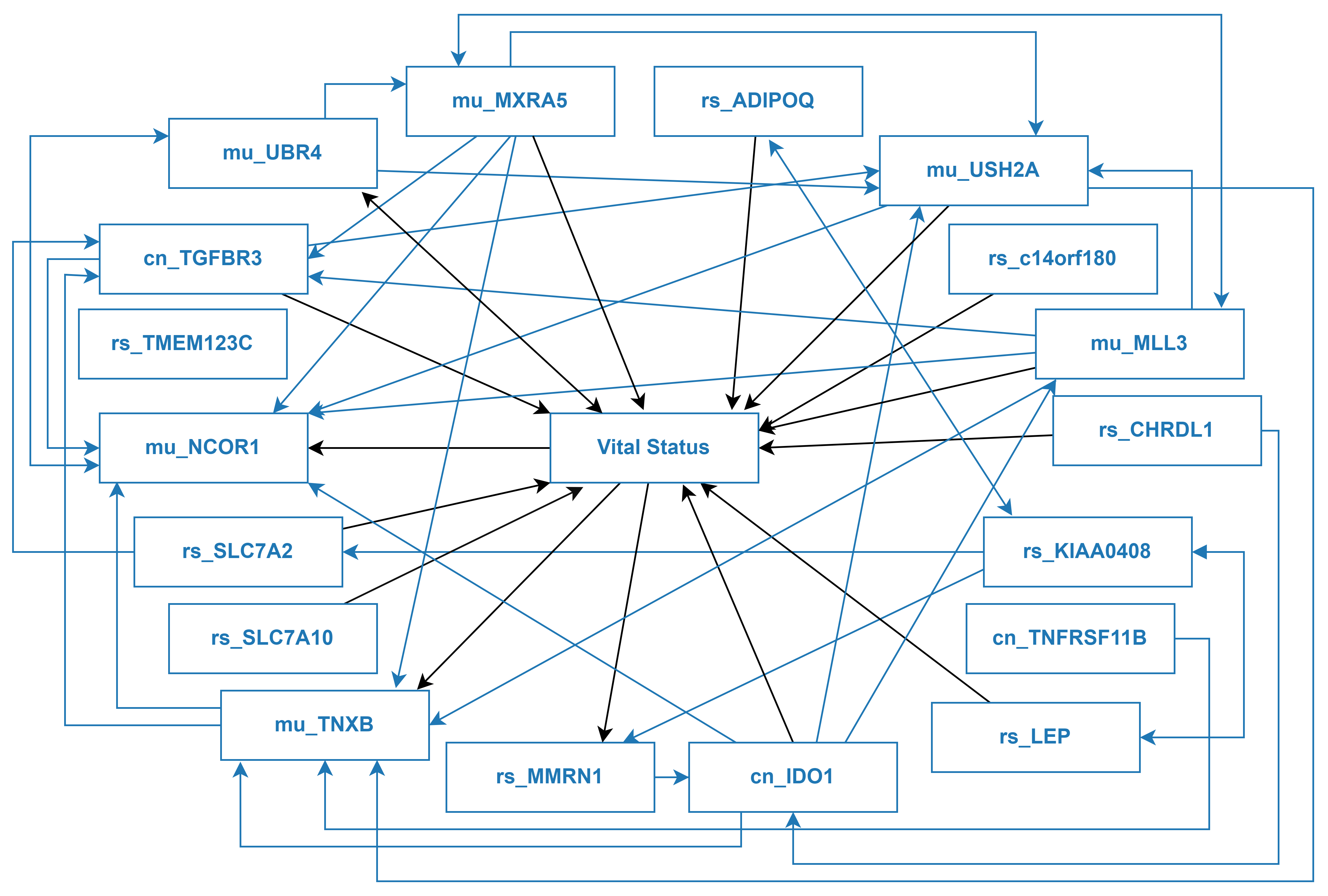}
    \caption{Graph of the PC algorithm applied on the Markov network produced by GPM  using a mixture of features selected from MMMB and MI feature selection.}
    \label{gpm}
\end{figure}

Given the nature of the dataset, and to conduct fair experimentation, both the continuous and the categorical data types had to be included. With the feature selection methods mentioned above, a subset of the data was obtained. Different statistical tests were conducted according to the type of data and the CD method implemented. CD methods produced directed graphs as is shown in Figures~\ref{fges_bic_mixed_mi}, \ref{ges_mmmmb_cv}, \ref{pc_cond_gauss_mi},
\ref{pc_chi_categorical}, and \ref{gpm}. The nodes in the graphs represent the features, for example, ``cn\_IDO1'' represents the copy number variations in gene IDO1. To demonstrate the efficacy of the CD methods, we backed our results from findings in the biomedical literature.


% \begin{figure}[ht]
%     \centering
%     \includegraphics[width=0.39\textwidth]{graphs_causal_discovery/ML703-GES_MMMB.drawio.png}
%     \caption{Graph of GES using the generalized cross-validation score on the categorical data from MMMB.}
%     \label{ges_mmmmb_cv}
% \end{figure}

\subsection{Using Categorical Data}
For categorical data, we used MMMB as the feature selection method to remove the independent variables given the Markov blanket of the target variable. The independence test between the variables was carried out by \texttt{testIndMultinom} in R. Then, we applied PC, FCI, and GES to the selected set of features. In the following sections, the results of PC and GES are described, while the results of FCI can be found in Appendix \ref{app-sect:CD}.

\subsubsection{Using PC}  
We applied the PC algorithm with the Chi-square test, which gave us the results shown in Figure~\ref{pc_chi_categorical}. From the causal graph, we found that vital status was connected to the mutation in genes MLL3 and TNXB and copy number variation in TNFRSF11B. Studies have shown that MLL3 is the $6^{th}$ most frequently mutated gene in ER+ breast cancer patients. In addition, it has been proven that mutations in MLL3 result in endocrine therapy resistance in patients, which can only be analyzed if the tumor samples of patients are genotyped \citep{stauffer2021mll3}. The other genes also play an important role in the survival of the patients. For example, the TNXB  gene, being a gene that is not mutated too frequently, is not discovered by mutation frequency-based software. However, the TNXB gene is crucial in the carcinogenesis of breast cancer \citep{10.1093/carcin/bgw065}. In terms of survival, the TNXB gene has been validated as a biomarker for early metastasis of breast cancer. Thus, it is beneficial that the CD methods can identify the factors affecting survival that are hard to obtain from the commonly used tools. Moreover, studies showed that the dysregulation of the TNFRSF11B gene results in poor prognosis of breast cancer patients \citep{luo2017dysregulation}, which ultimately results in distant organ metastasis of breast cancer. Therefore, this result backs the shorter median overall survival data of patients, which is also reflected in our results using the CD methods.

\begin{table}[]
    % \centering
    \caption{Prediction for masked tokens with BlueBERT and their respective scores. The masks are placed randomly in a sentence to evaluate the performance of the model.}
    \resizebox{\columnwidth}{!}{
    \begin{tabular}{llll}
    \toprule
    Claim                                                                                       & Masked Token &  \makecell[l]{Top Predicted Token} & Score \\
        \midrule
    \makecell[l]{{[}MASK{]} in gene UBR4 is related to the\\ survival in cancer }                                    & Mutation     & Mutation            & 0.773 \\
    \makecell[l]{Changes in protein levels in gene CDK1.PY15\\ is related to the {[}MASK{]} in cancer} & Survival     & Progression         & 0.135 \\
    
    \makecell[l]{Mutation in gene TNXB  is related to the\\ survival in {[}MASK{]} }                                  & Patients     & Patients            & 0.179 \\
    \makecell[l]{Copy number variations in gene COL14A1 is \\related to the survival in {[}MASK{]}}               & Cancer       & Cancer              & 0.299\\
        \bottomrule
    \end{tabular}}
    \label{tab:predd} 
\end{table}

% \begin{table}[t]
%   \centering 
%   \caption{Description with the main take-away point. Note that the caption should appear \emph{above} the table.}
%   \begin{tabular}{llll}
%   \toprule
%     \textbf{Claim} & \textbf{Masked Token} & \textbf {Top Predicted Token} & \textbf{Score} \\
%     \midrule
%    [MASK] in gene UBR4 is related to the survival in cancer & Mutation & Mutation & 0.773\\ 
%     Changes in protein-protein interactions in gene CDK1.PY15 is related to the [MASK] in cancer & Survival & Progression & 0.135\\ 
%     Mutation in gene TNXB  is related to the survival in [MASK] & Patients & Patients & 0.179 \\
%     Copy number variations in gene COL14A1 is related to the survival in [MASK] & Cancer & Cancer & 0.299\\
%     \bottomrule
%   \end{tabular}
%   \label{tab:example} 
% \end{table}

\subsubsection{Using GES} 
Lastly, we used the generalized CV score \citep{huang2018generalized} with the data obtained from MMMB on the GES method. From the graph shown in Figure~\ref{ges_mmmmb_cv}, we observed that the target variable (vital status) was related to many variables including mutations in genes UBR4, TNXB, MXRA5, USH2A, and MLL3. It was also connected to the copy number variations in genes IDO1, TNFRSF11B, and NCOR1. The role of genes TNXB, MLL3, and TNFRSF11B is mentioned above. Apart from that, MXRA5 plays a role in cancer cells in forming metastases hence affecting survival \citep{minafra2014gene}. IDO1 plays a role in the differentiation of monocytes, a type that has been found associated with tumor progression in breast cancer \citep{meireson2020ido}. For the scope of this paper, we will not be discussing other complex biological mechanisms and pathways for other genes, but there are studies suggesting the association of these genes with breast cancer and consequently overall survival.
% \begin{wraptable}{r}{6cm}
% \centering 
%   \caption{Performance of different models for the prediction of masked tokens in the claims produced by the causality models.}
%   \begin{tabular}{ll}
%   \toprule
%     \textbf{Model} & \textbf{Accuracy}\\
%     \midrule
%     DistilBERT & 56.25\%    \\
%    SciBERT & 68.75\%    \\
%    BlueBERT & \textbf{81.25}\%  \\
%     \bottomrule
%   \end{tabular}
%   \label{maskeee} 
% \end{wraptable}

% \begin{table}[!t]
% \caption{Performance of different models for the prediction of masked tokens in the claims produced by the causality models.\label{maskeee}}%
% \begin{tabular*}{\columnwidth}{@{\extracolsep\fill}ll@{\extracolsep\fill}}
% \toprule
% Model & Accuracy  \\
% \midrule
%  DistilBERT & 56.25\%    \\
%  SciBERT & 68.75\%    \\
%  BlueBERT & \textbf{81.25}\%  \\
% \botrule
% \end{tabular*}
% \end{table}

\subsection{Using Mixed Data}
For mixed data, we applied the MI feature selection method. The subset of data obtained included the 10 features that have the highest mutual information with the target variable (vital status). This subset was included to use different CD methods like FCI, FGES, PC, and GPM. We now show and discuss the results derived from the aforementioned CD methods. As mentioned, applying suitable causal discovery methods to the target variable (vital status) and its Markov blanket can find parents (direct causes) and children of the target variable \citep{Gao2015LocalCD}. All details regarding FCI can be found in Appendix \ref{app-sect:CD}.

% \subsubsection{Using FCI} 
% FCI was applied with the conditional Gaussian likelihood ratio test. The graph produced by FCI is shown in Figure~\ref{fci_mixed_mi_condgauss}. The vital status was found to be related to the gene expression level of gene SLC7A2. Research has propounded that increased levels of SLC7A2 in breast cancer patient samples are usually associated with poor overall prognosis and are considered a stand-alone variable for decreased survival, which stands with the results obtained from FCI. In addition to the relations between variables and the target variable, we observe the inter-variable relations as well. For the sake of brevity, we will only discuss the relationships between the variables connected with green lines in Figure~\ref{fci_mixed_mi_condgauss}, as these imply that the two variables connected do not contain any latent confounders. Therefore, the relations exhibited are true relations. This finding is backed by the research conducted in \cite{shi2017integrative} where TMEM132C and MMRN1 are highly associated with the stage of breast cancer as well as survival time.
\subsubsection{Using FGES} \label{sect:fges-mixed}
We applied FGES with the conditional Gaussian BIC score. It was observed that the target variable was related to the gene expression of the genes SLC7A2, SLC7A10, and MMRN1, as shown in Figure~\ref{fges_bic_mixed_mi}. Research has propounded that increased levels of SLC7A2 in breast cancer patient samples are usually associated with poor overall prognosis and are considered a stand-alone variable for decreased survival. Regarding SL7A10, it has been found to express less in breast cancer tissues compared to normal breast tissue. However, the relation with overall survival has not been investigated yet. MMRN1 has been recorded as a differentially expressed gene in many cancers and has the capability to be identified as a potential cancer biomarker. Moreover, MMRN1 expression has been observed to be related to the stage of breast cancer \citep{shi2017integrative}.
\subsubsection{Using PC}
Using the mixed data from MI feature selection, we implemented the PC algorithm with the conditional Gaussian likelihood ratio test. In this case, only one variable was observed to be related to the target variable, which was SLC7A2, as shown in Figure~\ref{pc_cond_gauss_mi}. The justification from the literature review was mentioned in the previous section.
\subsubsection{Using GPM}
Lastly, since GPM creates a Markov network using mixed data, we combined the features from MMMB (discrete) and the features from MI (continuous) along with the vital status as input to GPM. A Markov network was obtained using the adjacency matrix obtained from GPM. Then, an equivalence class was obtained using the PC algorithm (with RCIT) from the Markov network. Figure~\ref{gpm} shows the dense graph that is created using GPM. As can be observed, several variables have a relationship with the target variable, vital status, which show us the biological relations between the multi-omics features and a patient's survival. Validation of most of these variables from a biological perspective is mentioned in previous sections.

\subsection{Verification with Language Models}
The graphs produced by different CD methods showcase a wide variety of relationships between the target variable and other multi-omics data variables. Although we relied on the biomedical literature to validate our CD results, a more efficient approach was needed. The authenticity of these claims made by the causality model was validated by language models through different experiments. Various models were used for experimentation, but we only describe those that yielded the best results.

\subsubsection{Perplexity Score}
The perplexity of a language model measures the degree of uncertainty when it generates a new token, averaged over all words in the input. It has been used to do fact-checking \citep{lee-etal-2021-towards}, the idea being that claims that are supported by a given text corpus are expected to have a low perplexity, while those that are not supported would have a high perplexity.
If we have a claim made by the causality model, e.g, ``A is dependent on B'', we take its negation ``A is not dependent on B'', and we calculate the perplexity for both. The one with the lower perplexity is considered to be correct. We evaluated a set of claims using BlueBERT and SciBERT (using the same configuration and size as BERT-base). The results obtained conform with the claims from the causality models as shown in Table~\ref{perp}. To elaborate, the results from the CD models indicate that the mutation in gene UBR4 is associated with the survival status of cancer. In Table~\ref{perp}, the perplexity score for that particular claim with the best performing model (BlueBERT) is less, which represents the truthfulness of the statement.

\subsubsection{Masked Language Modeling}
One of the interesting uses of LLMs is that they can be used as knowledge bases \citep{lee-etal-2020-language,https://doi.org/10.48550/arxiv.1909.01066}. Using LLMs for such a purpose remains largely unexplored in the biomedical domain. We first used DistilBERT to assess its performance on the prediction of masked tokens from the claims. The claims from the causality model were masked randomly and fed into the model after they were tokenized. For every sentence, it gave us a computed probability score of different probable tokens for the masked token. As expected, the correct token predictions were less in number as compared to the incorrect predictions. To obtain good performance for this task, a model trained on biomedical data had to be used. Among the three models used for the prediction of masked tokens, BlueBERT produced the best results, with an accuracy of 81.25\%. DistilBERT and SciBERT lagged behind with the accuracy of predicting tokens at 56.25\% and 68.75\%, respectively. The performance of other models was lower compared to BlueBERT mainly because of the datasets they were trained on. We also show the performance of BlueBERT on various claims on different masks and most of the predicted masks were correct as we can see in Table~\ref{tab:predd}. The model was tested on a variety of claims made by CD models varying the placement of the masks. We also employed PubMedBERT \citep{DBLP:journals/corr/abs-2007-15779} trained on PubMed Abstracts and full-text articles from PubMedCentral and ClinicalBERT \citep{DBLP:journals/corr/abs-1904-05342} trained on clinical notes from MIMIC-III dataset \citep{johnson2016mimic} however, the scores for predicted masked tokens were lower than BlueBERT.

We also used BioGPT \citep{10.1093/bib/bbac409} to generate the relevant claims by providing a prompt. We observed the relevancy of the generated biomedical text produced by BioGPT when the input to the model consisted of a part of the claim. The generated text matched the claims made by the CD methods. Examples of generated text are in Appendix \ref{app-sect:Biogpt}.

% \section{Validity of the Results}\label{val_results}
% Using causal discovery for biomedical data certainly provides a different perspective on the problem. It can be an efficient approach to discovering various hidden changes in a patient's body. Causal discovery allows us to understand these relationships and to use the results for various downstream tasks, including the effect of newly discovered drug targets and the cause of a normal gene turning into a cancer gene. However, the reliability of the results from CD methods is questionable and the tolerance for incorrect results is particularly low in the biomedical domain. As a result, validation through research or professionals is required. In our paper, we chose to validate our findings using biomedical corpora. LLMs have largely helped the research body in several fields. The usage of language models for verifying the claims made by the causality model is efficient and reliable as it aids domain experts in further verification of the claims.
\section{Discussion}\label{discussion}
Using causal discovery for biomedical data certainly provides a different perspective on the problem. It can be an efficient approach to discovering various hidden changes in a patient's body. Causal discovery allows us to understand these relationships and to use the results for various downstream tasks, including the effect of newly discovered drug targets and the cause of a normal gene turning into a cancer gene. However, the reliability of the results from CD methods is questionable and the tolerance for incorrect results is particularly low in the biomedical domain. As a result, validation through research or professionals is required. In our paper, we chose to validate our findings using biomedical corpora. LLMs have largely helped the research body in several fields. The usage of language models for verifying the claims made by the causality model is efficient and reliable as it aids domain experts in further verification of the claims.

Our experimental results give an in-depth understanding of the different causal discovery methods used for multi-omics data. The data being mixed required us to delve deeper into the CD methods, their underlying assumptions (e.g., the faithfulness assumption for the PC algorithm), and technical assumptions (e.g., the linear-Gaussian assumption in the original GES for continuous data) in order for the results to be trustable. As can be deduced from our results, there are overlaps between the results of different methods. However, the differences suggest a need for increased consistency between the findings to assist decision-making in a medical setting. 
Overall, FGES and GES produced denser graphs compared to PC, but they are less efficient at handling mixed data in the current implementation. GPM helped in leveraging beneficial information from mixed data and modeling the distribution of our data which did not necessarily follow the extensively studied families of distributions. It adapts to our categorical variables containing different cardinalities. 

Causal graphs derived from different CD methods infer several important insights about the target variable. Out of all the graphs, the one produced by GPM followed by PC seems to be the best in terms of explainability from a medical lens. Vital status was found to be affected by several important variables. In some cases across the graphs, the vital status was observed to be the effect of a number of variables, for example, the mutations in gene MXRA5, and the cause for several specific variables such as the mutations in NCOR1. The latter observation, i.e., that vital status is the cause of changes in genes, might be unexpected at first glance. However, one example from the biomedical literature suggests that chemotherapy given to patients can sometimes result in alterations in the genome. Mutations in a patient's body can be related to resistance to therapy, dysregulation of cellular pathways, and metastasis.
%It can be attributed to the fact that the chemotherapy given to patients can sometimes result in alterations in the genome, as is propounded by the existing literature. Furthermore, mutations in a patient's body can also be related to resistance to therapy, dysregulation of cellular pathways, and metastasis.
% This observation, however, may prompt another thought process in which vital status is viewed as a reflection of the biological processes that occurred in a patient's body rather than a result of the events.  Thus, survival due to some reasons may cause the occurrence of other factors such as 

\section{Conclusion and Limitations}\label{conc}
Our work 1) exploited causal discovery approaches to unravel relationships between the survival of breast cancer patients with the multi-omics variables used in a subset of the TCGA dataset for breast cancer and 2) made use of language models for verification of the results discovered by CD. The CD approaches provide more interpretable ways to analyze data in the biomedical domain. Among the various CD methods implemented, we found that the GPM-based method yields the most comprehensive result. Due to the lack of methods that can validate the reliability of CD graphs, we leveraged the existing pre-trained language models to evaluate the claims made by the CD models. We conclude that models trained on relevant medical corpora like BlueBERT demonstrated superiority than other LLMs for the validation of biomedical claims. 

\paragraph{Limitations}
For CD models to be used for various complex applications, more efficient methods that can handle the mixed types of data are needed. Until now, the available software used for generating graphs is not scalable when a large dataset is used, which led us to reduce the number of features in our dataset drastically. This process might have affected the efficacy of our findings, especially since some related biological information can be considered crucial. In addition, more focus needs to be directed towards exploring validation methods for CD models. Furthermore, language models have been trained on data up to a certain point in time. For them to be used as a source of evaluation for a continuously evolving field of biology, they need to be trained on up-to-date biomedical data to ensure accurate evaluation. Moreover, one should use language models with caution and explore a variety of them for higher reliability, especially in the medical context.
% In future work, we aim to produce the whole causal graph over all the features that allow for latent common causes and cycles to produce a detailed understanding of the causal relations, although it is currently very challenging, given the current developments in the causal discovery field.

% \begin{table}[t]
%   \centering 
%   \caption{Description with the main take-away point. Note that the caption should appear \emph{above} the table.}
%   \begin{tabular}{llll}
%   \toprule
%     \textbf{Method} & \textbf{Metric1} & \textbf{Metric2} & \textbf{Metric3} \\
%     \midrule
%     Baseline & 1.1 & 2.3 & 0.1 \\ 
%     NetNet & 41.3 & 31.9 & 77.4 \\ 
%     \bottomrule
%   \end{tabular}
%   \label{tab:example} 
% \end{table}

% \begin{figure}[t]
%   \centering 
%   \includegraphics[width=2.5in]{plot.png} 
%   \caption{Description with the main take-away point. Note that figure captions should appear below the figure.}
%   \label{fig:example} 
% \end{figure} 

% ACKNOWLEDGEMENTS ONLY GO IN THE CAMERA-READY, NOT THE SUBMISSION
% \acks{Many thanks to all collaborators and funders!}
% \typeout{}
\clearpage
\bibliography{full-paper-template}
\clearpage
\appendix
\section*{Appendix}\label{app}

\section{Data Exploration} \label{app-sect:Data}
\begin{figure}[htbp]
    \centering
    \includegraphics[width=0.7\textwidth]{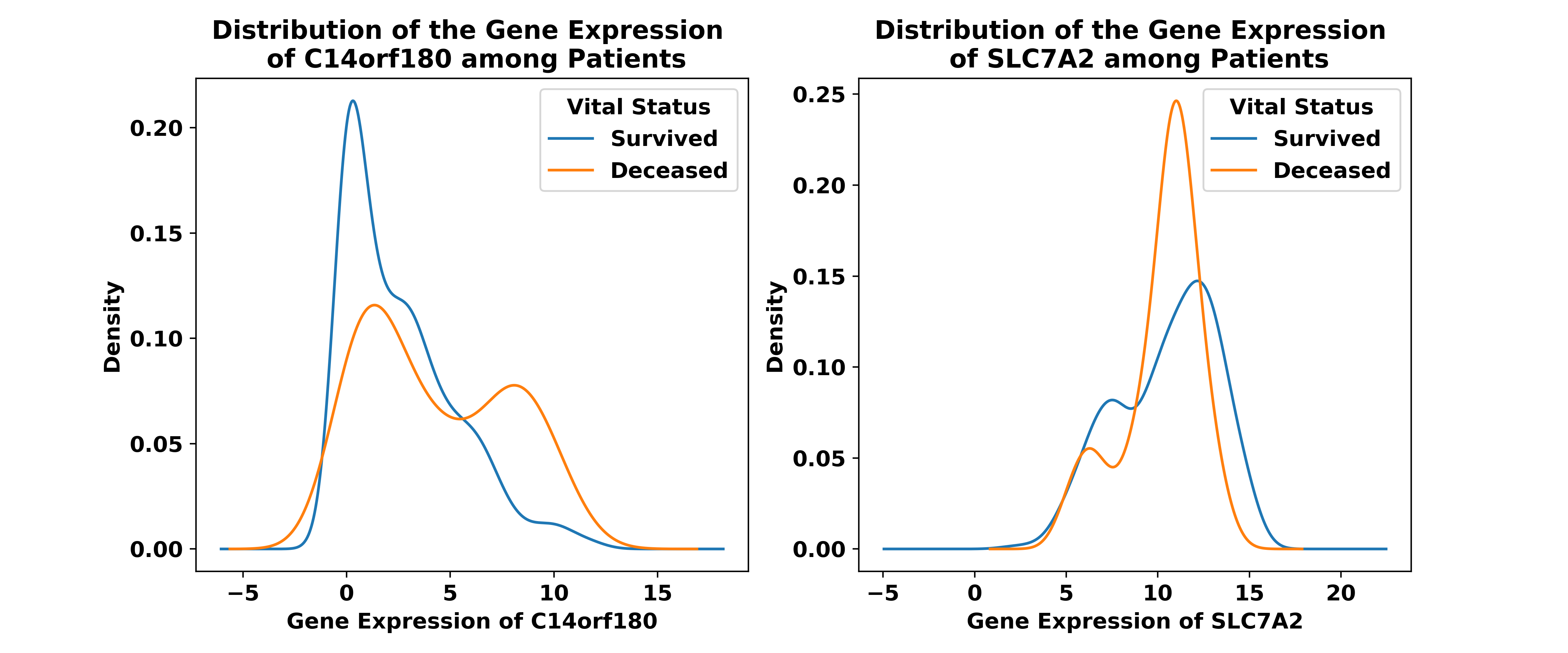}
    \caption{The distribution difference in gene expression across the two vital status categories of patients.}
    \label{fig:dist_rs}
\end{figure}

\begin{figure}
        \centering
        \includegraphics[width=1\textwidth]{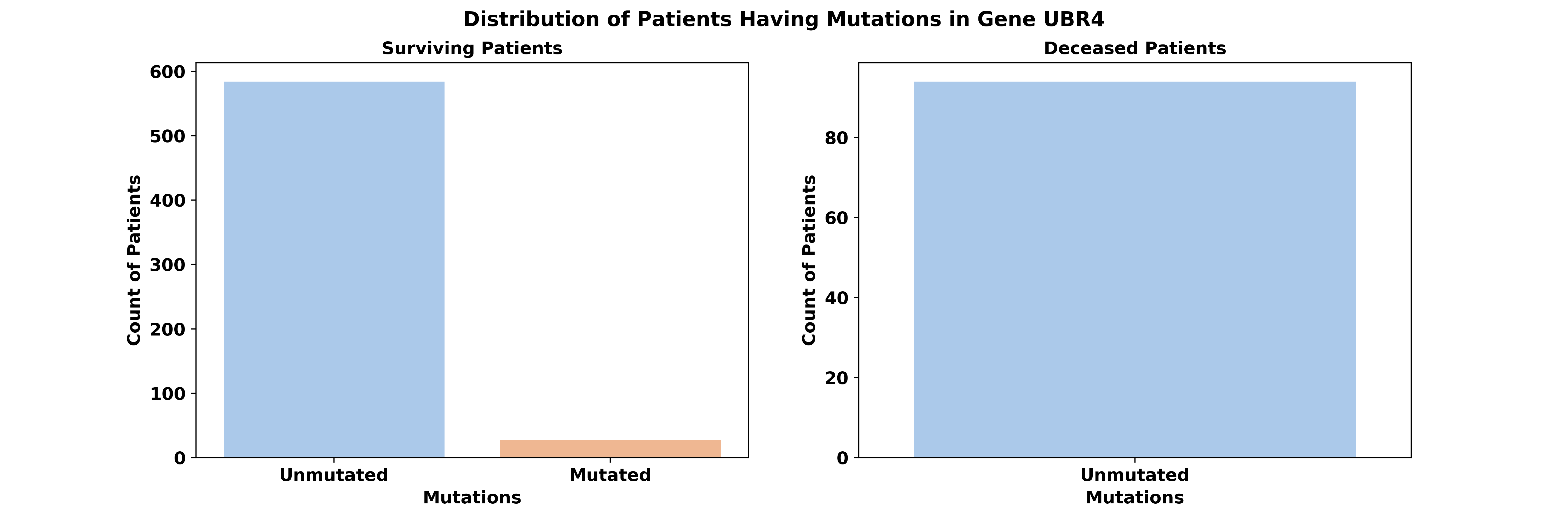}
        \caption{The patient distribution difference across the two vital status categories regarding the mutations present in gene URB4.}
        \label{fig:mu_dist_URB4}
\end{figure}

To better understand the samples in the data and different distributions of patients with varying alterations in the genome, we visualized the distributions as shown in the figures below. For mutations in gene MLL3, Figure~\ref{fig:mu_dist_MLL3} illustrates that for surviving patients, the percentage of mutations is much lower than it is for deceased patients. On the other hand, Figure~\ref{fig:mu_dist_URB4} shows that deceased patients did not have any mutations in gene URB4 while surviving patients did. Regarding the numerical variables, Figure~\ref{fig:dist_rs} shows the kernel density estimation of the gene expressions of two genes: C14orf180 and SLC7A2. The distribution of the gene expression levels in these genes also vary remarkably across surviving vs. deceased patients.\\
\begin{figure}[htbp]
         \centering
         \includegraphics[width=1\textwidth]{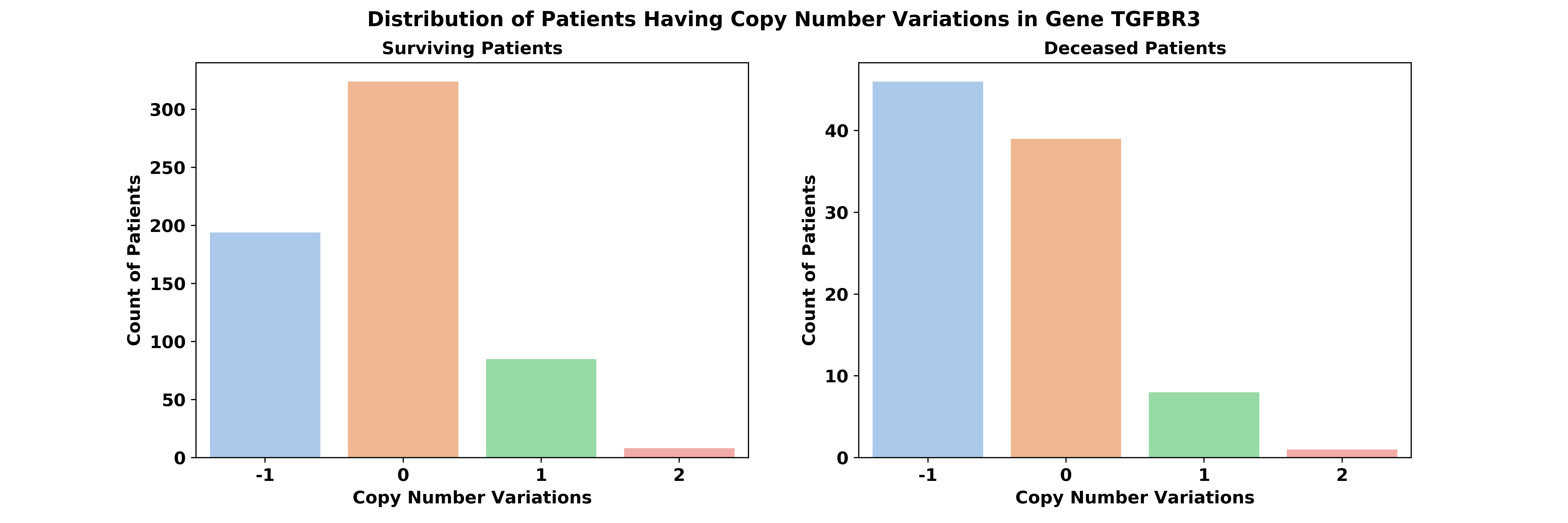}
         \caption{The patient distribution difference across the two vital status categories regarding the copy number variations in gene TGFBR3.}
         \label{fig:dist_cn_TGFBR3}
\end{figure}

\begin{figure}[htbp]
        \centering
        \includegraphics[width=1\textwidth]{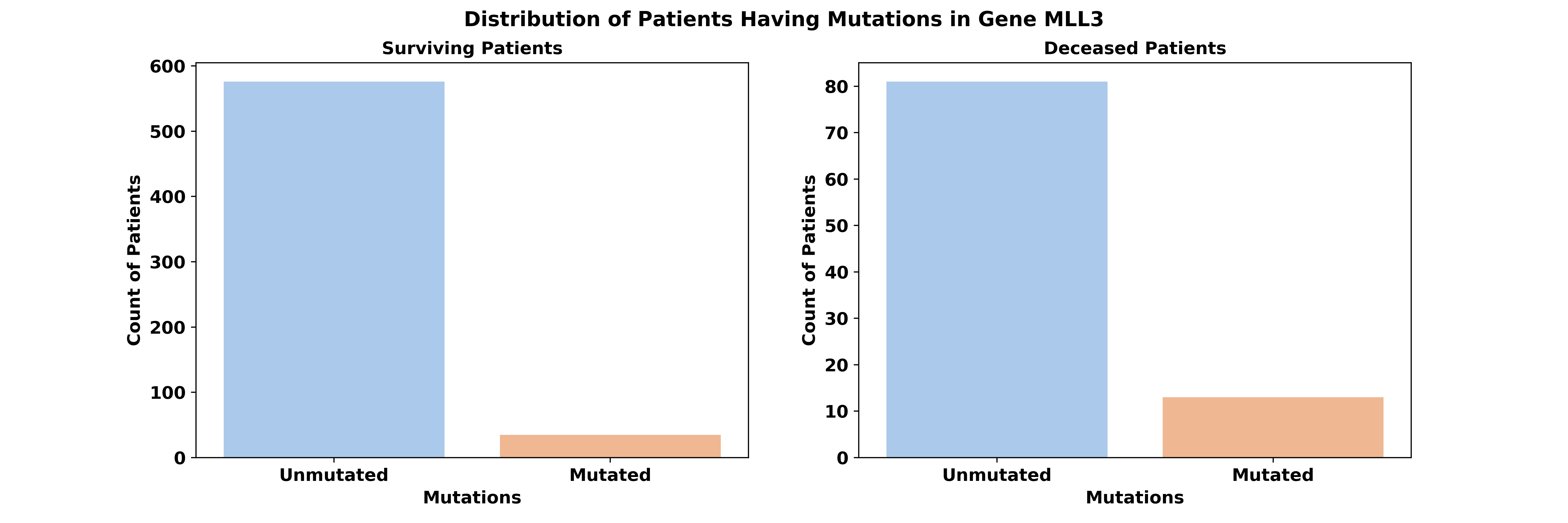}
        \caption{The patient distribution difference across the two vital status categories regarding the mutations present in gene MLL3.}
        \label{fig:mu_dist_MLL3}
\end{figure}

\clearpage
\section{Text Generation with BioGPT} \label{app-sect:Biogpt}
The text in Table~\ref{biogpt} was generated using different prompts incorporating variations in maximum tokens generated as well as the number of return sequences. Multiple prompts were used for a particular claim made by the causality models to analyze the difference in the generated text.
\begin{table}[htbp]
  \centering 
  \caption{Performance of different models for the prediction of masked tokens in the claims produced by the causality models.}
  \resizebox{\columnwidth}{!}{
  \begin{tabular}{ll}
  \toprule
    \textbf{Input Prompt} & \textbf{Generated Text}\\
    \midrule
    \makecell[l]{``Copy number variation in gene IDO1\\ affects vital status''} & \makecell[l]{ 1. Copy number variation in gene IDO1 affects vital \\status and therapeutic response \\in patients with multiple myeloma.}\\
                                                            & \makecell[l]{2. Copy number variation in gene IDO1 affects vital \\status in glioblastoma patients.}\\
                                                             & \makecell[l]{3. Copy number variation in gene IDO1 affects vital \\status of breast cancer cells.}\\
                                                            & \makecell[l]{4. Copy number variation in gene IDO1 affects vital \\ status in children with autism spectrum disorder.}\\
                                                            & \makecell[l]{5. Copy number variation in gene IDO1 affects vital \\ status and IDO1 expression is linked to disease onset\\  in type 1 diabetes.}   \\
                                                            \midrule
   \makecell[l]{``Change in gene expression in gene MMRN1 \\ affects vital status''} & \makecell[l]{1. Change in gene expression in gene MMRN1  \\ affects vital status of human breast cancer cell lines,\\ including hormone-dependent cells such as T47D.}\\
                                                                                  
                                                                                   &\makecell[l]{ 2. Change in gene expression in gene MMRN1 \\affects vital status of the offspring.}\\
                                                                                   &\makecell[l]{3. Change in gene expression in gene MMRN1 affects \\ vital status of the offspring.}\\
                                                                                  & \makecell[l]{4. Change in gene expression in gene MMRN1 affects \\ vital status of an aging human fibroblast line.}\\
                                                                                  & \makecell[l]{5. Change in gene expression in gene MMRN1 affects \\ vital status of an aging human fibroblast line.}\\
        
    \bottomrule
  \end{tabular}}
  \label{biogpt} 
\end{table}

\begin{figure}
    \centering
    \begin{minipage}[t]{0.47\textwidth}
        \includegraphics[width=\textwidth]{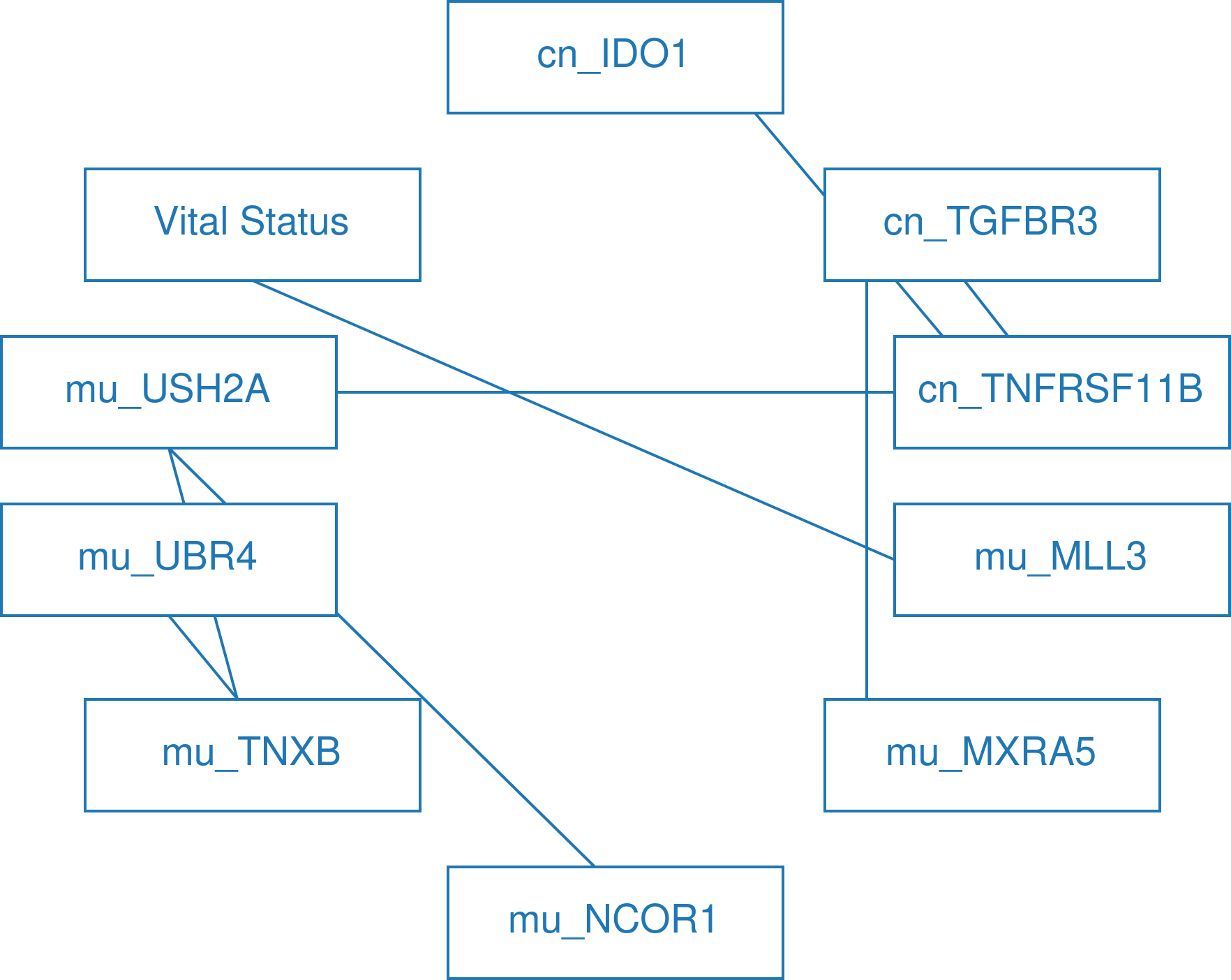}
        \caption{Graph of FGES using the BDeu score on the categorical data.}
        \label{Figure:fges_bdeu}   
   \end{minipage} 
\hfill
\centering
    \begin{minipage}[t]{0.49\textwidth}
            \includegraphics[width =\textwidth]{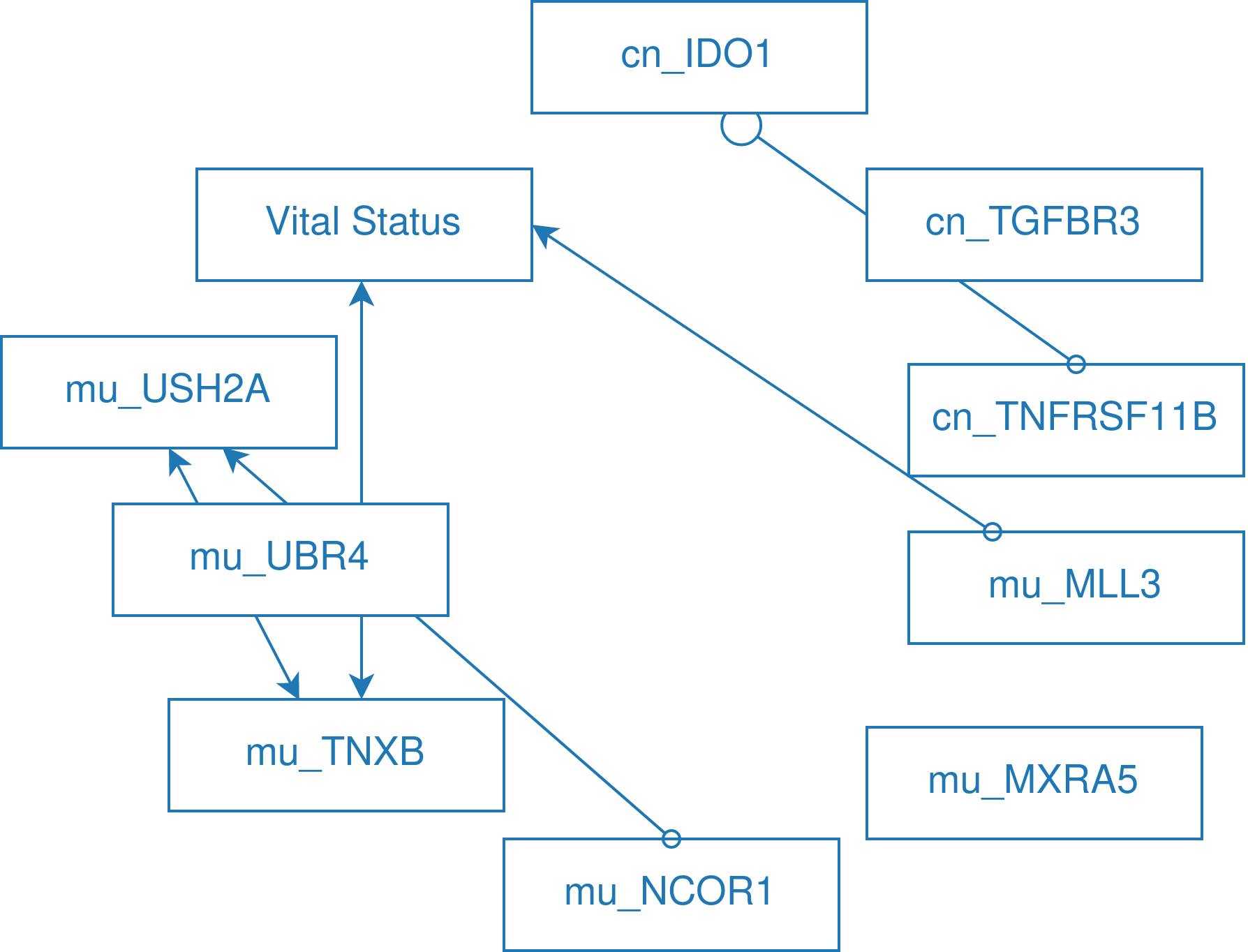}
    \caption{Graph of FCI using the Chi-square test on the categorical data.}
    \label{Figure:fci_chi}  
    \end{minipage}

\end{figure}

\section{Other Causal Discovery Methods} \label{app-sect:CD}
\subsection{FCI} 
FCI \citep{spirtes} is a generalization of the PC algorithm, which allows for latent confounders. Similar to PC, it starts with a completely undirected graph and forms an initial skeleton. All edges of the skeleton are of the form o-o, where \emph{o} can represent an arrow head or an arrow tail. Unlike PC, due to the latent variables, it is not enough to consider only subsets of neighborhoods of nodes $A$ and $B$ to decide whether the edge o-o for $A$ and $B$ should be removed. The bi-directed edges between nodes indicate the presence of an unmeasured confounder. To decide whether edge o–o should be removed, one computes Possible-D-SEP($A, B$) and Possible-D-SEP($B, A$) and performs conditional independence tests of $V_A$ and $V_B$ given all subsets of Possible-D-SEP$(x,y)$ and of Possible-D-SEP$(y, x)$. Then, the V-structures are formed \citep{spirtes}.

\subsection{Causality Results with Categorical Data}
\subsubsection{Using FGES}
The result of FGES with BDeu score using Tetrad is shown in Figure~\ref{Figure:fges_bdeu}. As can be observed, the target variable `vital status' was found to be related to the mutation in the gene MLL3. Matching our results, studies have shown that MLL3 is the 6th  most frequently mutated gene in ER+ breast cancer patients. In addition, it has been proved that mutations in MLL3 result in endocrine therapy resistance in patients, which can only be analyzed if the tumor samples of patients are genotyped. This can result in a better-structured treatment plan for patients. For the sake of experimentation, we used the discrete BIC score with FGES but it did not give any significant results.

\subsubsection{Using FCI}
Using the same subset obtained by MMMB, we applied FCI with the Chi-square test. The causal relationships obtained from this method were overlapping with the results we got previously, as shown in Figure~\ref{Figure:fci_chi}. This method also indicates the genes MLL3 and TNXB are related to the survival status or vital status of a breast cancer patient. The biological reasoning behind these relations has been already explored.
\begin{figure}[t]
    \centering
    \includegraphics[width=0.55\textwidth]{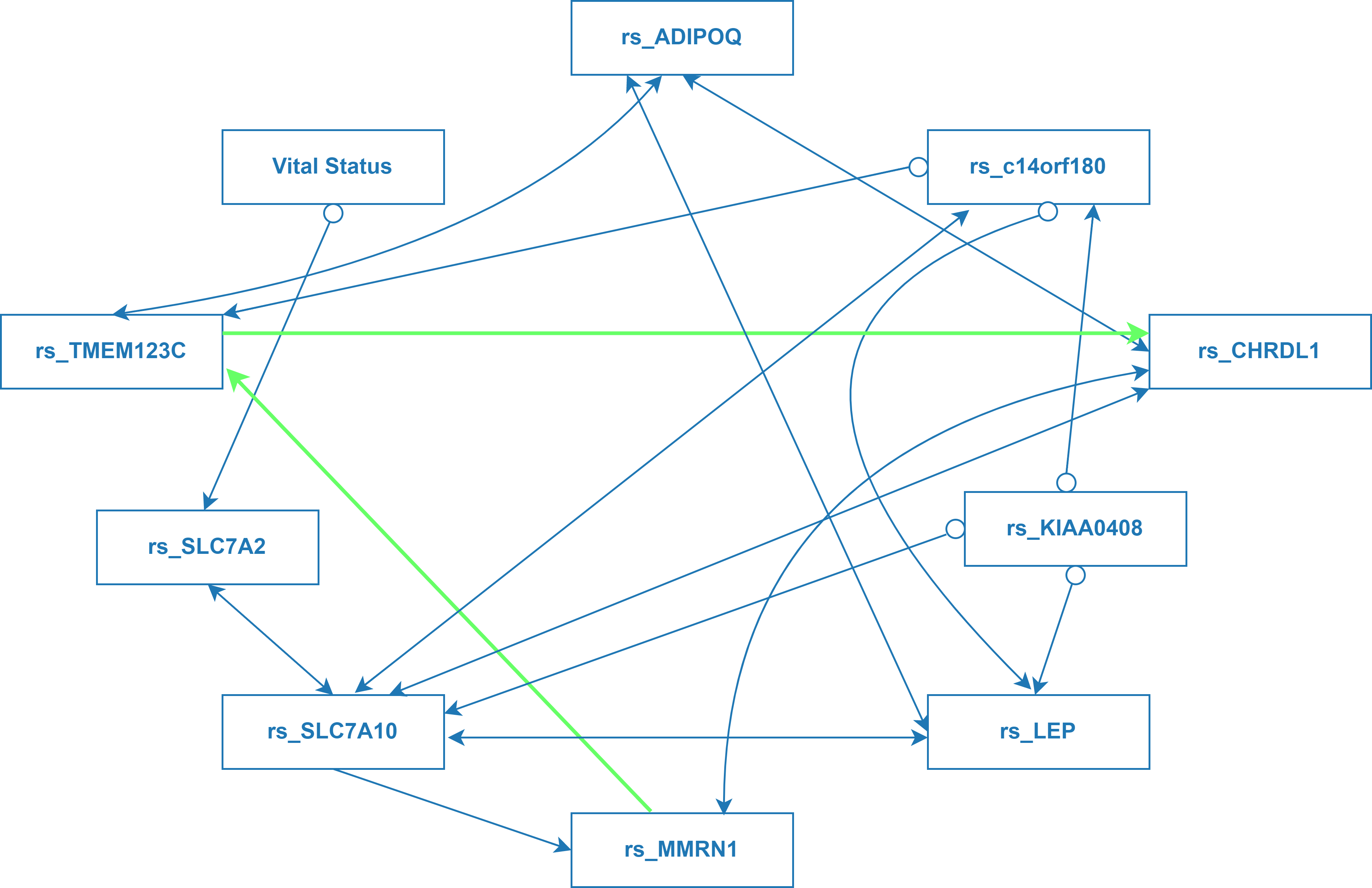}
    \caption{Graph of FCI using the conditional Gaussian likelihood ratio test on the mixed data from MI selection.}
    \label{fci_mixed_mi_condgauss}   

\end{figure}
\subsection{Causality Results with Mixed Data}
\subsubsection{Using FCI} 
FCI was applied with the conditional Gaussian likelihood ratio test. The graph produced by FCI is shown in Figure~\ref{fci_mixed_mi_condgauss}. The vital status was found to be related to the gene expression level of gene SLC7A2, the biological explanation is in Section \ref{sect:fges-mixed} and stands with the results obtained from FCI. In addition to the relations between variables and the target variable, we observe the inter-variable relations as well. For the sake of brevity, we will only discuss the relationships between the variables connected with green lines in Figure~\ref{fci_mixed_mi_condgauss}, as these imply that the two variables connected do not contain any latent confounders. Therefore, the relations exhibited are true relations. This finding is backed by the research conducted in \citep{shi2017integrative} where TMEM132C and MMRN1 are highly associated with the stage of breast cancer as well as survival time, reinforcing the usage of CD methods for clinical practice with proper supervision.

\end{document}